\documentclass[letterpaper, 10 pt, conference]{ieeeconf}  

\IEEEoverridecommandlockouts                              
\usepackage[utf8]{inputenc}
\usepackage[T1]{fontenc}
\usepackage{bm}
\usepackage{graphicx} 
\usepackage{amsmath}
\usepackage{algpseudocode}
\usepackage{algorithm}
\usepackage{xcolor}
\usepackage{algorithm}
\usepackage{algpseudocode}

\title{\LARGE \bf
Immersive Virtual Reality Platform for Robot-Assisted Antenatal \\Ultrasound Scanning
}

\author{Shyam A$^{1}$$^{,2}$ Aparna Purayath$^{2}$ Keerthivasan S$^{2}$ Akash S M$^{2}$ Aswathaman Govindaraju$^{1}$$^{,2}$ \\ Manojkumar Lakshmanan$^{2}$ and Mohanasankar Sivaprakasam$^{1}$$^{,2}$
\thanks{$^{1}$ Department of Electrical Engineering, Indian Institute of Technology Madras, India}%
\thanks{$^{2}$ Healthcare Technology Innovation Centre (HTIC),       Indian Institute of Technology Madras, India }%
\thanks{(e-mail: shyam@htic.iitm.ac.in)}
}

\begin{document}

\maketitle
\thispagestyle{empty}
\pagestyle{empty}

\begin{abstract}

Maternal health remains a pervasive challenge in developing and underdeveloped countries. Inadequate access to basic antenatal Ultrasound (US) examinations, limited resources such as primary health services and infrastructure, and lack of skilled healthcare professionals are the major concerns. To improve the quality of maternal care, robot-assisted antenatal US systems with teleoperable and autonomous capabilities were introduced. However, the existing teleoperation systems rely on standard video stream-based approaches that are constrained by limited immersion and scene awareness. Also, there is no prior work on autonomous antenatal robotic US systems that automate standardized scanning protocols. To that end, this paper introduces a novel Virtual Reality (VR) platform for robotic antenatal ultrasound, which enables sonologists to control a robotic arm over a wired network. The effectiveness of the system is enhanced by providing a reconstructed 3D view of the environment and immersing the user in a VR space. Also, the system facilitates a better understanding of the anatomical surfaces to perform pragmatic scans using 3D models. Further, the proposed robotic system also has autonomous capabilities; under the supervision of the sonologist, it can perform the standard six-step approach for obstetric US scanning recommended by the ISUOG. Using a 23-week fetal phantom, the proposed system was demonstrated to technology and academia experts at MEDICA 2022 as a part of the KUKA Innovation Award. The positive feedback from them supports the feasibility of the system. It also gave an insight into the improvisations to be carried out to make it a clinically viable system.  
\end{abstract}

\section{INTRODUCTION}

Maternal mortality is one of the widely accepted key indicators of a country's health and socioeconomic development~\cite{c_1}. It is often higher in rural settings than urban areas due to inadequate access and unaffordable healthcare. Also, the availability of skilled healthcare professionals and the access to health resources~\cite{c2}, like primary health services, medicines, infrastructure, etc, are limited. The World Health Organisation's (WHO) Antenatal Care (ANC) model recommends eight ANC contacts during the period of pregnancy~\cite{c3}. Early and regular pregnancy scans can detect the majority of fetal structural defects (59\%), chromosomal defects (78\%)~\cite{c4} and improve the overall maternal care management. 

\begin{figure}[!htb]
    \centering
    \includegraphics[width=0.5\textwidth]{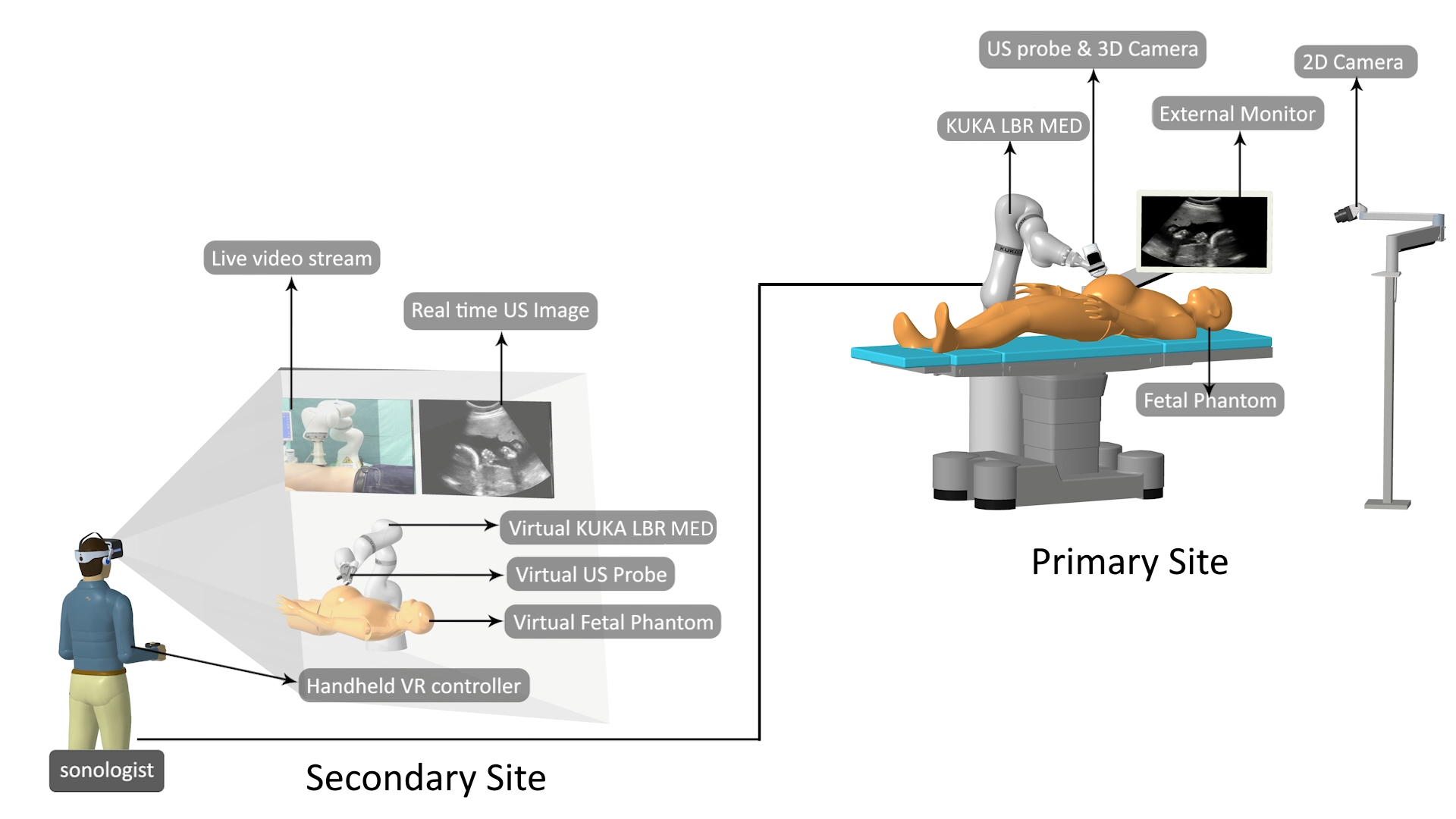}
    \caption{Proposed system architecture}
    \label{SystemArch4}
\end{figure}

Access to quality maternal and fetal care can be enhanced by equipping health centers with robotic ultrasound systems. Antenatal robotic ultrasound technology is the fusion of US imaging and robotics for non-invasive fetal imaging during pregnancy. Systems with teleoperation, collaborative assistance, and autonomous capabilities at varied levels of robot autonomy (LORA)~\cite{c5} exist. These robotic systems allow for more precise and consistent imaging~\cite{c6}, standardized scanning, and improve comfort and safety for patients as well as sonologists~\cite{c7}. Further, telemedicine and teleconsultation provide remote medical consultations in rural areas. The comparative studies of teleoperated US  imaging from Arbeille et al.~\cite{c8} and Xie et al.~\cite{c9} have suggest that US-based remote diagnosis is as effective and useful as manual interventions.

Research and clinical studies on robotic fetal ultrasonography are limited. iFIND - intelligent Fetal Imaging and Diagnosis system~\cite{c10} aims at automating ultrasound fetal examinations. It follows a customized workflow to scan the desired anatomical location in a consistent way. The robotic US acquisition follows a generic path that is not specific to any scan pattern prescribed for antenatal scanning, like the six-step approach recommended by the International Society of US in Obstetrics and Gynecology (ISUOG)~\cite{c11}.

Tsumura et al.~\cite{c12} and Arbeille et al.~\cite{c13} proposed teleoperated robotic systems for fetal scanning. The majority of such implementations of robot-assisted remote US systems use an audio-visual channel for examination. However, these standard approaches lack a sufficient degree of immersion and scene awareness~\cite{c14}. Although VR technology can address these shortcomings, it has not been implemented before. The current research on VR for medicine mostly focuses on surgical training, psychiatric treatment, pain management, and rehabilitation~\cite{c15} but not on antenatal ultrasound scanning.

A novel platform to address the shortcoming mentioned above is proposed in this work. As shown in Fig. 1, it combines the use of robotics with VR technology for antenatal US examinations.
The significant contributions in that regard are:

1. An immersive virtual reality platform for the sonologist to control the robotic arm over a wired network is developed. It provides an enhanced visual representation of the clinical setting, including the robot and patient's anatomy, and offers haptic feedback-based robotic manipulation, resulting in a more realistic experience. Additionally, real-time US acquisition and streaming allows for instant and accurate diagnosis. 

2. An autonomous robotic system, which automates the ISUOG's six-step approach for obstetric US scanning is developed. These standardized scans are autonomously performed by the robot under the supervision of the sonologist, who can observe the robotic movements through the VR headset and command the course of the probe at any point of time.

This paper is organized as follows: Section II provides an overview of the system, including its components and communication methods. Section III describes the design and development of manual contact and autonomous modes. Section IV presents the observations related to the demonstration of the proposed system on a fetal phantom. Lastly, Section V has the conclusions and future work.

\begin{figure*}[!htb]
    \centering
    \includegraphics[width=\textwidth]{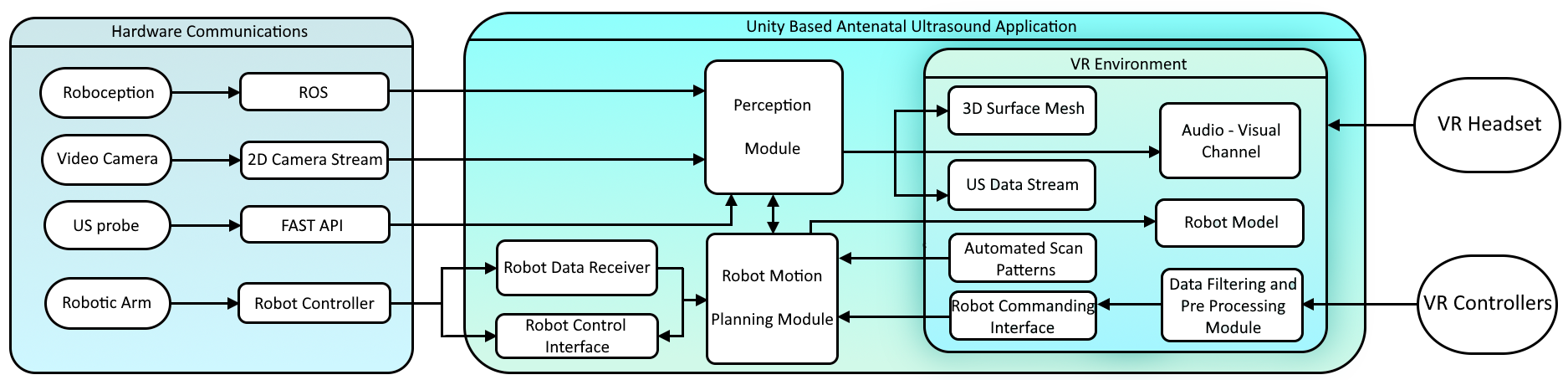}
    \caption{Schematic of communication between different components of the system}
    \label{Intro}
\end{figure*}

\section{SYSTEM OVERVIEW}

\subsection{System Components}


The system comprises a primary and a secondary site. The primary site consists of a 7 Degrees Of Freedom (DOF) KUKA LBR Med robot arm attached with two end effectors: a 3D stereo camera and a curvilinear US  probe. Additionally, a 2D camera has been integrated into the system to enable real-time patient interaction. The secondary site, operated primarily by a sonologist, features an Oculus VR headset to provide an immersive user interface and enable the robot to be steered manually or autonomously using the VR controllers. The primary and secondary sites were connected via a wired network. A Unity-based VR application (shown in Fig. 2) was developed to provide Graphical User Interface (GUI) and facilitate communication between the system components. An oval-shaped abdomen US phantom with a 23-week fetus was used for the preliminary trials.

\subsection{Robot Communication}
The communication channel between the robot and the VR application primarily uses the Fast Robot Interface (FRI). As depicted in Fig. 2, through the FRI’s data read channel, the Robot Data Receiver fetches the robot’s current status (joint and cartesian values, error status, etc.) in real-time at a rate of 500 Hz. The Robot Control Interface uses the FRI’s write channel to command and overlay the robot’s motion. A Java application is deployed and externally controlled from the VR application over a TCP/IP network. It encloses and commands state changes in the FRI connection.

\begin{figure*}[!htb]
    \centering
    \includegraphics[width=\textwidth]{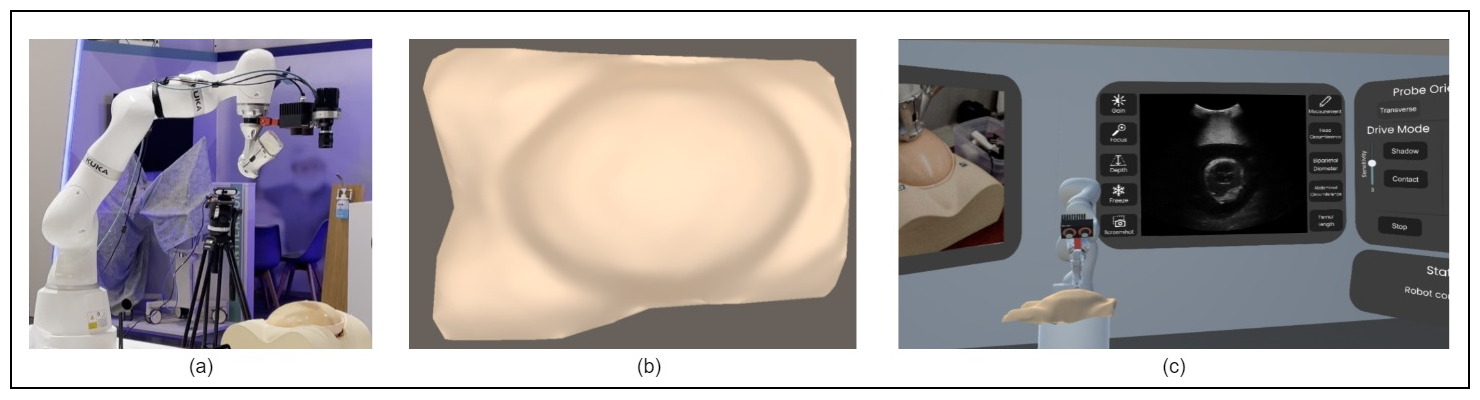}
    \caption{(a) Robot initialization (b) Reconstructed phantom anatomy (c) VR environment }
    \label{Intro1}
\end{figure*} 

\subsection{Interfacing 3D Camera}
The system utilizes a stereo camera from Roboception (rc visard 65 monochrome) equipped with a pattern projector to reconstruct the patient's anatomy. Communication with the camera was established using the Robotic Operating System (ROS) via the GenICam interface for seamless data transfer. In addition, the ROS bridge interface is utilized to enable effective communication between ROS and the Unity software for transferring data. 

\subsection{Real-time Ultrasound Streaming}
FAST (Framework For Heterogeneous Medical Image Computing And Visualization) interface was used to stream live US images at 30 fps from a US sensor - Clarius C3 HD. The US sensor uses Wi-Fi Direct for streaming data to the application.

\subsection{Immersive Virtual Reality Environment}
The VR space offers an immersive and enhanced visual experience that enables sonologists to improve patient care quality. The robot model is represented using the Unified Robotics Description Format (URDF) inside the virtual environment. The URDF file contains a range of kinematic and dynamic parameters, including linear and angular friction, damping, and stiffness. Thus an accurate representation of the robot's physical behavior is simulated. The reconstructed patient anatomy is loaded as a mesh file into the VR space, as shown in Fig. 3b, and its coordinates are mapped to the robot's base frame. The user interface dashboard consists of three segments: the first segment streams live video from the patient site. The second segment streams real-time US, with the option to tune imaging parameters, such as gain, depth, and brightness. The third segment is drive mode selection, which allows the sonologist to switch between manual contact mode and autonomous mode for distinct scan patterns. The US probe orientations - longitudinal or transverse, can also be selected from this segment.

\section{METHODOLOGIES}

\subsection{Anatomical Surface Reconstruction }

As shown in Fig. 3a, the robot is initialized to a configuration that facilitates the 3D camera to have adequate coverage of the site of phantom placement. Next, the position of the phantom is adjusted to ensure that all the ArUco markers are in the vicinity of the 3D camera. Multiple perspectives of the phantom are captured as Point Cloud Data (PCD) using the 3D camera. The outliers in the acquired PCD are filtered using RANdom SAmple Consensus (RANSAC). Then, the filtered point cloud data is merged as a single PCD using the Iterative Closest Point (ICP) technique~\cite{c16}. For visualization purposes, a mesh is reconstructed from the PCD using the Poisson Surface Reconstruction algorithm~\cite{c17}.

\subsection{Manual Contact Mode}
The manual mode enables the sonologist to manipulate and control the robot in real-time via a wired network. The key feature of this mode is that it enables the US probe to maintain contact with the patient's anatomy throughout the scan, thereby ensuring good-quality US imaging. By utilizing position control and force monitoring, the robot can maintain a permissible contact force at the end effector (i.e., the US probe) while maneuvering. 

To achieve real-time control of the robot, the hand gestures of the sonologist are captured using the constellated IR LEDs within the VR controller. These movements are read as position and orientation data in VR space using the Unity software. An inherent coordinate mapping is constructed from the VR to the robot space. This mapping allows the representation of the VR controller's position and orientation values in robot space. As a preprocessing step, a sequence of filtering algorithms is applied to these values to prevent unintended robot motions. Initially, the position and orientation values are given as input to a workspace filter to validate whether those values are within the robot's dexterous workspace~\cite{c18}. This workspace was determined by limiting the probe's orientation to a 60-degree cone arc~\cite{c19}. Post the workspace filter, acceleration, and velocity filters are implemented to prevent jerks. The linear and angular velocity parameters are limited to 20 mm/s and 30 deg/s, respectively. The normalized and spherical lerping methods are used to create smooth transitions between the filtered poses.   

\emph{Postion Control}:  PCD provides an excellent geometric approximation of the patient's anatomy. In order to accurately determine the mapping of camera coordinates in the robot's frame of reference, a Hand-Eye Calibration~\cite{c20} was performed between them. This mapping allows to represent the PCD in the robot space. The filtered position values are superimposed on the PCD. The vertical components of the position values (Z axis) are updated to match the PCD contour. This ensures that the cartesian position of the robot is confined to the contour of the PCD. Any variations along the vertical component (up and down movements) from the VR controller will not reflect in robot motion. 

It is crucial to ensure that the robot's motion avoids reaching any singular configurations. The rank of the Jacobian matrix was continuously verified to detect singularities. Since the system uses a redundant manipulator, the pseudoinverse of the Jacobian matrix needs to be computed, and it is expressed as:

\begin{equation} \label{eu_eqn}
J^+ = J^T {[J J^T]}^{-1},
\end{equation}

where $J$ represents the Jacobian matrix, and $J^+$ represents the Moore-Penrose inverse. 

\emph{Force Monitoring}: The US probe used in the current system lacks force-sensing capabilities at the contact point. Hence, the contact forces were monitored using the robot's joint torque sensors. Using the model of the robot dynamics, the joint torques are converted into end-effector forces. As a safety measure, both the resultant force of the end effector $(F_r)$ and its component $(F_s)$ along the probe axis are continuously monitored to ensure that they remain within the minimum $(F_c)$ and maximum $(F_m)$ permissible values, i.e., 
\begin{equation} \label{eu_eqn1}
F_c < (F_r,F_s) < F_m
\end{equation}

Also, the force $F_s$ along the probe axis is used for monitoring the contact with the anatomy.

By combining position control and force monitoring, the robot is made to traverse the probe along the anatomy contour and maintaining the permissible contact force. Thereby allowing the sonologist to capture US images without causing discomfort to the patient.

\subsection{Autonomous Mode}
Autonomous robotic US systems mitigate the repetitive nature of standard procedures for sonologists by automating US scans, thus providing an efficient and consistent solution to streamline the diagnostic process. In the case of antenatal US scanning, ISUOG recommends a standard six-step approach for determining various fetal parameters during the second and third trimesters~\cite{c11}. These steps include determining the fetal presentation, detecting fetal cardiac activity, identifying the number of fetuses in the uterus, determining the location and position of the placenta, estimating amniotic fluid, and measuring fetal biometrics such as the Biparietal Diameter (BPD), Head Circumference (HC), Abdominal Circumference (AC), and Femur Length (FL). ISUOG also specifies the recommended US probe scanning position and orientation on the anatomy to determine each parameter. These scanning patterns are well-standardized and have become a regular part of the sonologist's examination routine.
\begin{figure}[!htb]
    \centering
    \includegraphics[width=0.475\textwidth]{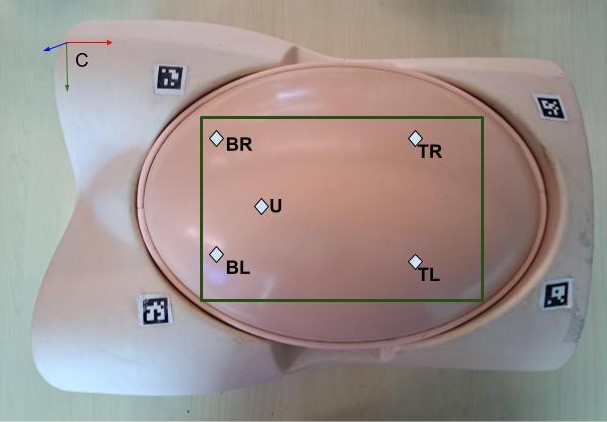} 
    \caption{5 key points computed on the fetal phantom}
    \label{SystemArch1}
\end{figure}

\begin{algorithm}
\caption{Path Finding Algorithm}\label{alg:cap}
\begin{algorithmic}
\Require $\mathbf{p}_s, \mathbf{p}_e,$ \texttt{PCD\_data,} $sd$\\
$\mathbf{v}_{se} = \mathbf{p}_e - \mathbf{p}_s$  \\
$m_{s_e} = ||\mathbf{v}_{se}||_{2}$  \\
$n = \frac{m_{s_e}}{sd}$ \Comment{$sd \rightarrow$ minimum distance between two points}\\

$\mathbf{P} = [\ ]\  $\\ 
$\mathbf{N} = [\ ]\  $ 
\State $i \gets 0$  
\While{\texttt{$i < n$}}
    \If{$i = 0 $} \Comment{$pz_{i-1}$ $\rightarrow $ vertical component of $\mathbf{p}_{i-1}$}
        \State $pz_{i-1} \gets 0$ 
    \EndIf   \Comment{unit vector $\hat{\mathbf{v}}_{se}$ }
    \State \texttt{$\mathbf{pSeudo}_i \gets  (\mathbf{p}_s + ({\hat{\mathbf{v}}_{se}}* sd *i)) + \left[0,0,pz_{i-1}\right]$} 
    \State \texttt{$\mathbf{p}_i \gets \texttt{NN}(\mathbf{pSeudo}_i,\texttt{PCD\_data})$}
    \State \texttt{$\mathbf{P}.\texttt{append}(\mathbf{p}_i)$}
    \State \texttt{$\mathbf{n}_{zi} = \textbf{Normalz}(\mathbf{p}_i)$}
\State $i \gets i + 1$    
\EndWhile\\

$\mathbf{P} = \left[ \mathbf{p}_0, \mathbf{p}_1, \dots, \mathbf{p}_{n-1} \right] $  \Comment{path points from PCD}\\
$\mathbf{N} = \left[ \mathbf{n}_{z0}, \mathbf{n}_{z1},....., \mathbf{n}_{zn-1} \right] $ \Comment{point normals  from PCD}\\

$\mathbf{pathPoints} = \texttt{PolyFIT}(\mathbf{P},sd)     $\\
$\mathbf{pathNormals}   =  \texttt{SmoothenZ}(\mathbf{N})$                 
\end{algorithmic}
\end{algorithm}
 
The developed system assists sonologists by automating these scans. Like manual contact mode, the system uses position and force monitoring to maintain skin contact and autonomously scan the segment. All these scans can be interpolated as geometric patterns using 5 key points, namely, the Umbilicus point (U), Bottom Left (BL), Bottom Right (BR),  Top Left (TL), and Top Right (TR). The Umbilicus (anatomical landmark) has to be manually selected by the sonologist. The application has the provision to choose the other key points manually, or it can be geometrically computed using the Umbilicus point and the ArUco markers. Any scanning pattern can be approximated to lines and curves using these key positions. Fig. 4 illustrates the position of all 5 key points computed for the fetus phantom.

Each pattern's probe positions and orientations are computed using a path planning algorithm, i.e., Algorithm 1. The path planner is defined by path points $\mathbf{P}$ and normals $\mathbf{N}$. A directional vector $\mathbf{v}_{se}$ is formed from the starting point $\mathbf{p}_s$ pointing towards the ending point $\mathbf{p}_e$ on the PCD. The vector $\mathbf{v}_{se}$ is discretized into $n$ pseudo-points $(\mathbf{pSeudo}_i)$ based on sampling distance $sd$. A KDtree search algorithm, denoted by $\texttt{NN}$, is used to find the closest points to the pseudo points on the PCD. These points are connected to form a smooth path using polynomial fitting methods. The probe's orientation is calculated based on the normal vector of each path point and the scan type (longitudinal or transverse) using the axis-angle formulation. The desired positions and orientations of the probe are transformed into the robot's space using the established coordinate mapping. The linear velocities are obtained by numerical differentiation of the position values. The space-fixed angular velocities are derived from the orientations using the expression $\dot{R}R^{T}$, where R is the rotation matrix corresponding to the robot's current orientation.

Finally, the Jacobian matrix of the robot is used to map the obtained task-space velocities ($\dot{X}$) to joint velocities ($\Dot{{\Theta}}$), using the relation $\Dot{\Theta} = J^+\dot{X}$.

\section{OBSERVATIONS AND DISCUSSIONS}

\subsection{Manual Control Mode}

\begin{figure}[!htb]
    \centering
    \includegraphics[width=0.475\textwidth]{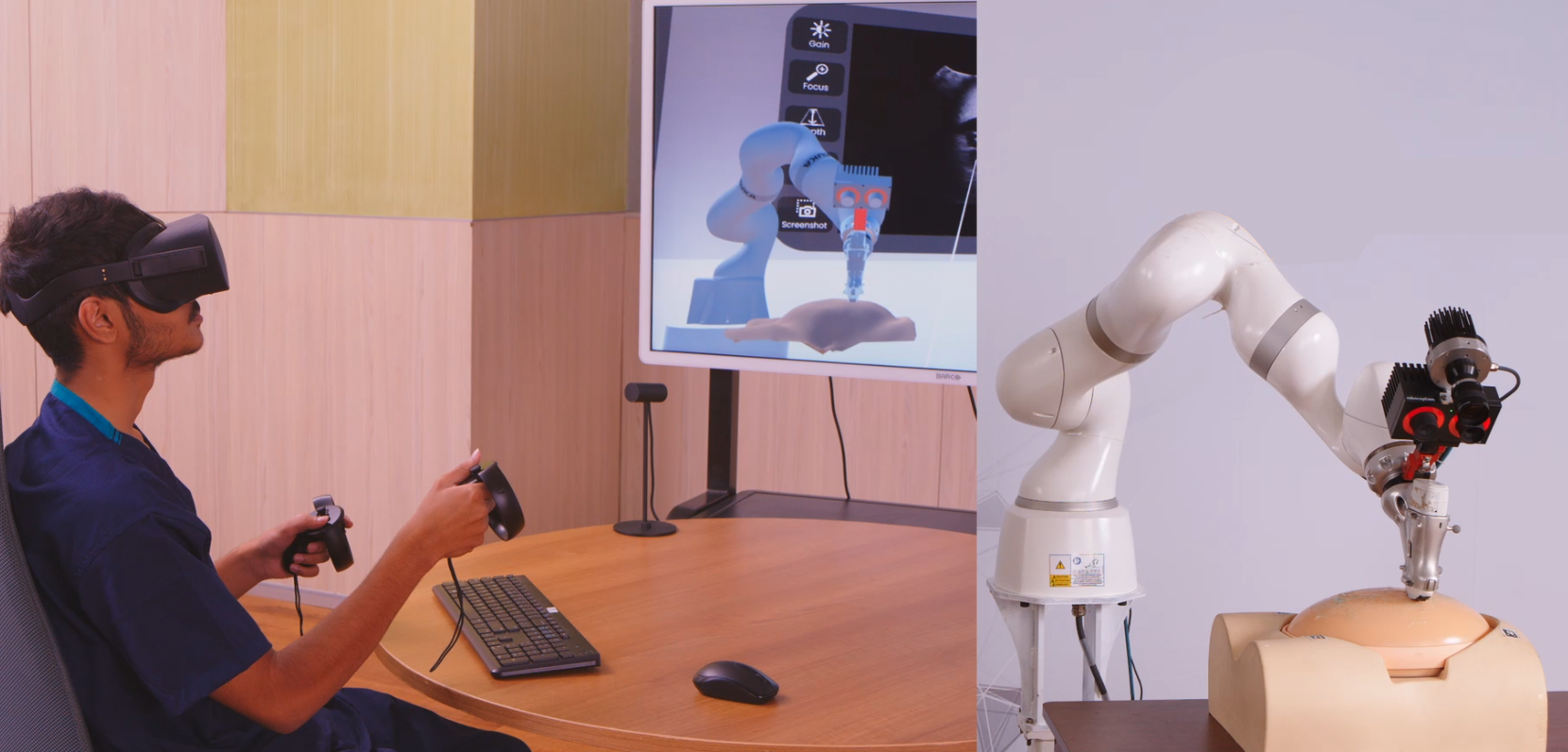} 
    \caption{Demonstration of manual control mode}
    \label{SystemArch3}
\end{figure}

The present study demonstrates the ability to exercise real-time control of a robotic arm through a wired network, as shown in Fig. 5. To ensure a stable connection, the system continuously monitors jitter and packet loss. The robot is only maneuvered when the latency is within the range of 5 to 8 ms. The effectiveness of the manual contact mode heavily relies on the transfer of rigid body motion from the VR controller to the robotic arm. As shown in Fig. 6, the algorithm has eliminated the high-speed variations and accidental drops of the VR controller. During these disturbances, the robot's pose stays intact and prevents unintended motions. For phantom demonstration, the minimum $(F_c)$ and maximum $(F_m)$ permissible forces required to maintain skin contact were set to 2N and 5N, respectively. The haptic feedback is given to the VR controller based on the variations in the robot's joint forces and position values. A high vibration alert is given to the user when the interaction forces are closer to $(F_m)$.    

\begin{figure}[!htb]
    \centering
    \includegraphics[width=0.475\textwidth]{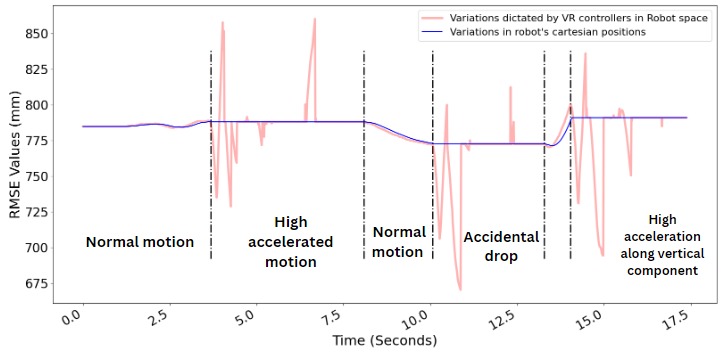} 
    \caption{VR controller input Vs Robot cartesian movement }
    \label{SystemArch2}
\end{figure}

The developed system was demonstrated at MEDICA 2022, and more than 50 participants volunteered to experiment with the system. They were provided with a rudimentary demonstration of the working model. Without any mention of its safety features, participants were asked to use the system. The users were able to actuate the robot along all 6 DOF, involving only translatory, rotary, or simultaneous translatory and rotary movements along the three independent axes. The system exhibited the capability to eliminate all types of disturbances, including workspace limitations and singular configurations. No adverse incidents of VR sickness were reported by any of the participants. However, some individuals experienced a minor degree of discomfort after using the VR for approximately 25 minutes. 

The proposed system can be easily extended to a telerobotic platform, provided the connectivity is facilitated through a high-speed internet network. The prospective advancements entail the implementation of telerobotic manipulation with due consideration given to network latency, bandwidth, and security, which are known to pose significant technical challenges.

\subsection{Autonomous Mode}

\begin{figure}[!htb]
    \centering
    \includegraphics[width=0.375\textwidth]{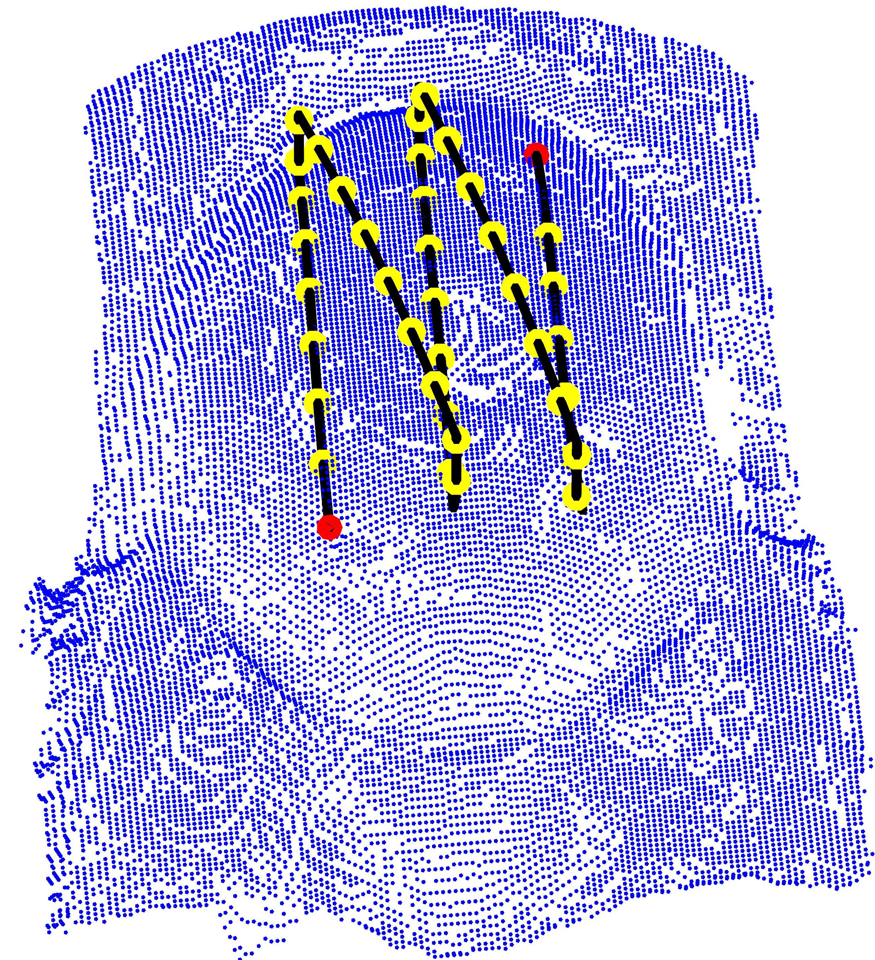} 
    \caption{Overlay of the computed path on the PCD to scan and identify the number of fetuses}
    \label{SystemArch}
\end{figure}
The developed autonomous system is classified as LORA-5, where the automation provides a predetermined set of options, and the human operator must select one for the system to carry out. In our current setup, once the Umbilical point (U) has been selected, the system computes the path and corresponding orientation of the probe for each scan pattern. The sonologist is provided with the choice to select any scanning patterns from the user dashboard, and the autonomous robot motion is initiated. For instance, Fig. 7 displays one such computed path on PCD to implement the number of fetuses scan pattern. The position control and force monitoring ensures the contact between the anatomy and the US probe by maintaining the interaction forces between $F_c$ and $F_m$ values.
\begin{figure}[!htb]
    \centering
    \includegraphics[width=0.475\textwidth]{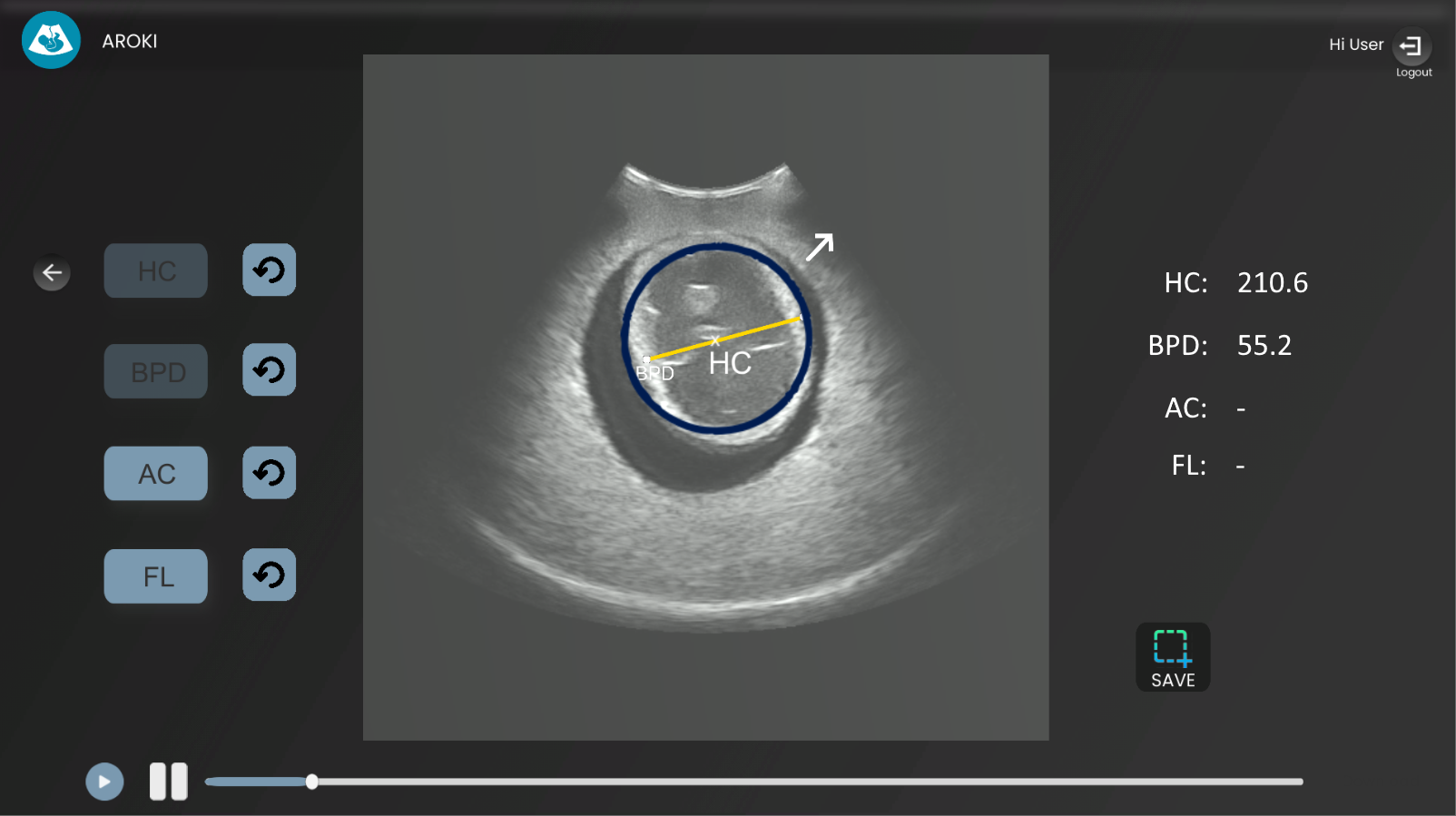} 
    \caption{Fetal measurements for the phantom}
    \label{Measurement1}
\end{figure}

The system allows the user to switch between autonomous and manual contact modes for diagnosis. Additionally, it includes a feature that enables the sonologist to pause the robot's motion and annotate fetal measurements. For example, the US image obtained during the autonomous scan of the fetus phantom and the measurements annotated by a sonologist at MEDICA are shown in Fig. 8. The system also records the streams of US images, which can be utilized for post-analysis or expert review.

\section{CONCLUSIONS}

This paper presents a new system designed for robot-assisted antenatal scanning using an immersive VR platform. Manual contact mode through a wired network and autonomous mode adapted to the standard six-step approach are used interchangeably in this system. The integration of VR, robotics, and the US in the proposed system enhances the sonologist's perception and experience of the patient environment. In addition, one potential application of VR in fetal monitoring is in training healthcare professionals. It provides a safe and controlled environment to practice and improve skills with a minimal learning curve during the transition from training to real-world scenarios. Another advantage is that the supervised autonomous feature of the system, specialized to the clinically relevant ISUOG scanning protocol, helps the sonologist reduce the time and effort spent on performing these routine scans on all patients. The system was successfully demonstrated at MEDICA 2022 using a 23-week fetal phantom, and the resulting observations are reported in this paper.  However, the system's usability and performance need to be comprehensively validated with clinical metrics. The real-world clinical environment poses a significant challenge in achieving seamless communication for telerobotics over a secure network and in addressing the unpredictable fetal movements during autonomous scans. The future scope is to achieve telerobotics and to autonomously manipulate the robot by leveraging US image feedback to compensate for fetal movements. We envisage this technology to be further extended as a surgical diagnostic and interventional platform that can address the lack of skilled resources and infrastructure.

\addtolength{\textheight}{-12cm}   




\section*{ACKNOWLEDGMENT}
We would like to thank KUKA AG, Germany, for giving us the opportunity to integrate their robotic platform to develop this system. The authors would like to acknowledge Dr. TejaKrishna Mamidi for his assistance in editing the manuscript.

\bibliographystyle{ieeetr}
\bibliography{main}

\begin{thebibliography}{10}

\bibitem{c_1}
J.~R. Wilmoth, N.~Mizoguchi, M.~Z. Oestergaard, L.~Say, C.~D. Mathers,
  S.~Zureick-Brown, M.~Inoue, and D.~Chou, ``A {N}ew {M}ethod for {D}eriving
  {G}lobal {E}stimates of {M}aternal {M}ortality,'' {\em Statistics,
  {P}olitics, and {P}olicy}, vol.~3, no.~2, 2012.

\bibitem{c2}
T.~Girum and A.~Wasie, ``Correlates of maternal mortality in developing
  countries: an ecological study in 82 countries,'' {\em Maternal {H}ealth,
  {N}eonatology, and {P}erinatology}, vol.~3, p.~19, 2017.

\bibitem{c3}
W.~H. Organization {\em et~al.}, ``{WHO} recommendations on antenatal care for
  a positive pregnancy experience: summary: highlights and key messages from
  the world health organization's 2016 global recommendations for routine
  antenatal care,'' tech. rep., World Health Organization, 2018.

\bibitem{c4}
B.~Whitlow, I.~Chatzipapas, M.~Lazanakis, R.~Kadir, and D.~Economides, ``The
  value of sonography in early pregnancy for the detection of fetal
  abnormalities in an unselected population,'' {\em British International
  Journal of Obstetrics \& Gynaecology}, vol.~106, no.~9, pp.~929--936, 1999.

\bibitem{c5}
F.~{V}on Haxthausen, S.~B{\"o}ttger, D.~Wulff, J.~Hagenah,
  V.~Garc{\'\i}a-V{\'a}zquez, and S.~Ipsen, ``Medical {R}obotics for
  {U}ltrasound {I}maging: {C}urrent {S}ystems and {F}uture {T}rends,'' {\em
  Current {R}obotics {R}eports}, vol.~2, pp.~55--71, 2021.

\bibitem{c6}
J.~M. Beer, A.~D. Fisk, and W.~A. Rogers, ``Toward a framework for levels of
  robot autonomy in human-robot interaction,'' {\em Journal of {H}uman-robot
  {I}nteraction}, vol.~3, no.~2, pp.~74--99, 2014.

\bibitem{c7}
M.~Bucolo, G.~Bucolo, A.~Buscarino, A.~Fiumara, L.~Fortuna, and S.~Gagliano,
  ``Remote {U}ltrasound {S}can {P}rocedures with {M}edical {R}obots: {T}owards
  {N}ew {P}erspectives between {M}edicine and {E}ngineering,'' {\em Applied
  Bionics and Biomechanics}, vol.~2022.

\bibitem{c8}
P.~Arbeille, R.~Provost, K.~Zuj, D.~Dimouro, and M.~Georgescu, ``Teles-operated
  echocardiography using a robotic arm and an internet connection,'' {\em
  Ultrasound in Medicine \& Biology}, vol.~40, no.~10, pp.~2521--2529, 2014.

\bibitem{c9}
W.~Xie, P.~Dai, Y.~Qin, M.~Wu, B.~Yang, and X.~Yu, ``Effectiveness of
  telemedicine for pregnant women with gestational diabetes mellitus: an
  updated meta-analysis of 32 randomized controlled trials with trial
  sequential analysis,'' {\em BMC {P}regnancy and {C}hildbirth}, vol.~20,
  no.~1, pp.~1--14, 2020.

\bibitem{c10}
J.~Housden, S.~Wang, X.~Bao, J.~Zheng, E.~Skelton, J.~Matthew, Y.~Noh,
  O.~Eltiraifi, A.~Singh, D.~Singh, {\em et~al.}, ``Towards {S}tandardized
  {A}cquisition with a {D}ual-{P}robe {U}ltrasound {R}obot for {F}etal
  {I}maging,'' {\em IEEE {R}obotics and {A}utomation {L}etters}, vol.~6, no.~2,
  pp.~1059--1065, 2021.

\bibitem{c11}
A.~Abuhamad, Y.~Zhao, S.~Abuhamad, E.~Sinkovskaya, R.~Rao, C.~Kanaan, and
  L.~Platt, ``Standardized {S}ix-{S}tep {A}pproach to the {P}erformance of the
  {F}ocused {B}asic {O}bstetric {U}ltrasound {E}xamination,'' {\em American
  {J}ournal of {P}erinatology}, vol.~33, no.~01, pp.~090--098, 2016.

\bibitem{c12}
R.~Tsumura and H.~Iwata, ``Robotic fetal ultrasonography platform with a
  passive scan mechanism,'' {\em International Journal of Computer Assisted
  Radiology and Surgery}, vol.~15, no.~08, pp.~1323--1333, 2020.

\bibitem{c13}
P.~Arbeille, J.~Ruiz, P.~Herve, M.~Chevillot, G.~Poisson, and F.~Perrotin,
  ``Fetal tele-echography using a robotic arm and a satellite link,'' {\em
  Ultrasound in Obstetrics and Gynecology}, vol.~26, no.~3, pp.~221--226, 2005.

\bibitem{c14}
P.~Stotko, S.~Krumpen, M.~Schwarz, C.~Lenz, S.~Behnke, R.~Klein, and
  M.~Weinmann, ``A {VR} system for {I}mmersive {T}eleoperation and {L}ive
  {E}xploration with a {M}obile {R}obot,'' in {\em 2019 IEEE/RSJ International
  Conference on Intelligent Robots and Systems (IROS)}, pp.~3630--3637, IEEE,
  2019.

\bibitem{c15}
L.~Li, F.~Yu, D.~Shi, J.~Shi, Z.~Tian, J.~Yang, X.~Wang, and Q.~Jiang,
  ``Application of virtual reality technology in clinical medicine,'' {\em
  American {J}ournal of {T}ranslational {R}esearch}, vol.~9, no.~9,
  pp.~3867--3880, 2017.

\bibitem{c16}
J.~Zhang, Y.~Yao, and B.~Deng, ``Fast and {R}obust {I}terative {C}losest
  {P}oint,'' {\em IEEE Transactions on Pattern Analysis and Machine
  Intelligence}, vol.~44, no.~7, pp.~3450--3466, 2021.

\bibitem{c17}
M.~Kazhdan, M.~Bolitho, and H.~Hoppe, ``Poisson surface reconstruction,'' SGP
  '06, (Goslar, DEU), p.~61–70, Eurographics Association, 2006.

\bibitem{c18}
K.~Mathiassen, J.~E. Fjellin, K.~Glette, P.~K. Hol, and O.~J. Elle, ``An
  {U}ltrasound {R}obotic {S}ystem {U}sing the {C}ommercial {R}obot {UR5},''
  {\em Frontiers in Robotics and AI}, vol.~3, pp.~1--16, 2016.

\bibitem{c19}
A.~Linc{\'e}, X.~Capelle, S.~Lepage, F.~Kridelka, and C.~Van~Linthout, ``Impact
  of the angle used in 2d ultrasonography on the foetal femur diaphysis
  measurement.,'' {\em Facts, Views \& Vision in Obgyn}, vol.~9, no.~2, p.~101,
  2017.

\bibitem{c20}
K.~H. Strobl and G.~Hirzinger, ``Optimal hand-eye calibration,'' in {\em 2006
  IEEE/RSJ {I}nternational {C}onference on {I}ntelligent {R}obots and
  {S}ystems}, pp.~4647--4653, IEEE, 2006.

\end{thebibliography}

\end{document}